\documentclass[
twocolumn,
]{ceurart}

\usepackage{listings}
\begin{document}

\copyrightyear{2024}
\copyrightclause{Copyright for this paper by its authors.
  Use permitted under Creative Commons License Attribution 4.0
  International (CC BY 4.0).}

\conference{Accepted in the IEEE Global Engineering Education Conference 2024 (EDUCON24), Kos, Greece}

\title{Knowledge Graphs as Context Sources for LLM-Based Explanations of Learning Recommendations}

\tnotemark[1]
\tnotetext[1]{You can use this document as the template for preparing your
  publication. We recommend using the latest version of the ceurart style.}

\author[1]{Hasan A. Rasheed}[%
orcid=0000-0002-2921-4809,
email=hasan.abu.rasheed@uni-siegen.de,
]
\cormark[1]

\address[1]{University of Siegen, Siegen, Germany}

\author[1]{Christian Weber}[%
orcid=0000-0001-6606-5577,
email=christian.weber@uni-siegen.de
]

\author[1]{Madjid Fathi}[%
orcid=0000-0002-7602-9593
]

\cortext[1]{Corresponding author.}

\begin{abstract}
  In the era of personalized education, the provision of comprehensible explanations for learning recommendations is of a great value to enhance the learner's understanding and engagement with the recommended learning content. Large language models (LLMs) and generative AI in general have recently opened new doors for generating human-like explanations, for and along learning recommendations. However, their precision is still far away from acceptable in a sensitive field like education. To harness the abilities of LLMs, while still ensuring a high level of precision towards the intent of the learners, this paper proposes an approach to utilize knowledge graphs (KG) as a source of factual context, for LLM prompts, reducing the risk of model hallucinations, and safeguarding against wrong or imprecise information, while maintaining an application-intended learning context. We utilize the semantic relations in the knowledge graph to offer curated knowledge about learning recommendations. With domain-experts in the loop, we design the explanation as a textual template, which is filled and completed by the LLM. Domain experts were integrated in the prompt engineering phase as part of a study, to ensure that explanations include information that is relevant to the learner. We evaluate our approach quantitatively using Rouge-N and Rouge-L measures, as well as qualitatively with experts and learners. Our results show an enhanced recall and precision of the generated explanations compared to those generated solely by the GPT model, with a greatly reduced risk of generating imprecise information in the final learning explanation.
\end{abstract}


\maketitle

\section{Introduction}

In personalized education, the provision of precise and comprehensible explanations for learning recommendations is both important and challenging. Explainability of learning recommendations has been found to enhance the student acceptance of the recommended content \cite{ooge_explaining_2022}. However, the precision of the textual and visual explanations faces several limitations when these are generated semi- or full-automatically. This is due to the level of knowledge, understanding, and reflection, that the learner needs from the generated explanation, which an automatic system may not be capable to offer. The recent research on LLMs opened new possibilities for solutions to generate explanations for learning recommendations on a deeper linguistic and semantic level. The massive datasets used to train such LLMs, and the high number of features considered by the model, enables generating more informative texts for the explainability task. However, they face serious concerns about their ability to generate a sufficiently precise text to reflect the learning intent from the recommendation, especially in sensitive domains. On the one hand, the technology readiness level (TRL) of the majority of LLMs still does not exceed the level TRL-2 \cite{yan_practical_2023}. From an ethical perspective, the transparency of LLMs is rarely more than Tier-1 \cite{yan_practical_2023}, based on the classification of transparency by Chaudhry et al. \cite{rodrigo_transparency_2022}. This apprehension aligns with the observations of Fullan et al. \cite{fullan_artificial_2023}, who also expressed concerns about various adverse effects of ChatGPT in education. They highlighted issues such as the lack of originality in its responses, potentially leading to answers that lack meaning and fail to stimulate exploration or imagination due to their linearity or flatness \cite{dana_chatgpt_2023}. Despite these drawbacks, Hargreaves \cite{hargreaves_words_2023} underscored that utilizing such technology in education exhibits promise, advocating against its outright rejection.

Our research addresses this challenge by combining the strengths of KGs and Generative Pre-Trained Transformers (GPT) models, aiming to enhance the precision, and thus reliability, of the explanations they generate for learning recommendations. We utilize the features of KGs to enhance the querying process of LLMs, through extracting contextual knowledge from the KG and then performing an informed prompt engineering process that is supported by pedagogy experts’ inputs. The resulting LLM prompt is then used to guide the model away from generating wrong or irrelevant information and limit the scope of answers it provides to the curated information provided from the knowledge graph. This ensures more precise answers, while still utilizing the model’s ability to combine pieces of information and phrase a human-like explanation, which learners can easily comprehend

\section{Background}

Knowledge Graphs are structured representations of knowledge, composed of entities and the relationships between them \cite{fensel_introduction_2020}. They serve as powerful tools for organizing information, capturing semantic connections, and providing a foundation for contextual understanding. A KG is defined as in (1). In educational contexts, KGs can be constructed from curated educational materials, forming a reliable source of factual knowledge \cite{abu-rasheed_building_2023}.

\begin{equation}
 KG = {(h,r,t) | h,t \in E , r \in R},
\end{equation}

Where h is the head entity, t is the tale entity, and r is the relation between h and t. E is the entity group and R is the relation group.

The use of KGs to support LLMs is gaining an increased focus recently \cite{pan_unifying_2023, luo_reasoning_2023, logan_baracks_2019}, which can be traced back to: 1) the new breakthroughs of LLMs, 2) the growing concerns about their risks and limitations \cite{yan_practical_2023}, and 3) the effective role KGs can play in reducing LLM hallucinations and enhancing their accuracy \cite{sequeda_benchmark_2023}. Sequeda et al. show that KGs-based querying enhanced the results of a one-shot question answering task in comparison to SQL-querying, when a KG representation of the SQL database was used \cite{sequeda_benchmark_2023}.
KGs have also been well utilized for explainability tasks in the literature. This is due to their potential to provide causal reasoning \cite{munch_interactive_2019}, factual knowledge \cite{logan_baracks_2019}, and interpretable semantic relations \cite{luo_reasoning_2023, abu-rasheed_transferrable_2022, rajabi_knowledge-graph-based_2022}. To that end, KGs not only enhance the reasoning of the LLM with facts, but also offer more information, which influences the precision of the automatically generated text. This information is parsed to the LLM in form of a prompt context that is designed during the prompt engineering phase. Depending on the LLM used, contextual parts of the prompt act as guidelines that steer the model to generate more relevant outputs. We build on the concepts in \cite{abu-rasheed_transferrable_2022} and \cite{pan_unifying_2023} to utilize KG’s structure and semantic relations for enriching the contextual part of OpenAI’s GPT-4 model prompt \cite{openai_gpt-4_2023}, thus enhancing the precision of its explanations of a learning recommendation.

\section{Methodology}
To utilize the GPT-4 potentials while reducing the amount of irrelevant and imprecise text it generates for the learning recommendation, we propose a strategy for constructing the data structures and the LLM prompt to maximize the amount of factual and contextual information that the model receives before generating an explanation for the recommendation, see Fig. 1. Contextual information that KG provides is acquired from the structural relations and the metadata of the learning content connected to the recommended elements. Both types of information are then translated into textual strings that are inserted as prompt context in the LLM query.

\subsection{Knowledge Graph Structure}
We create the KG from educational materials, aligning with a pre-defined taxonomy, to ensure comprehensive coverage of relevant topics. Our taxonomy includes four levels: 1) Learning goals, 2) Courses, 3) Topics, and 4) Open educational resources (OER). For simplicity, we use the term learning object (LO) as defined in \cite{noauthor_ieee_2020} hereafter to refer to any of the last three taxonomy levels, unless they are explicitly named. Learning goals and LOs are represented by graph nodes, who`s properties include their titles and descriptions among other metadata. We focus on textual properties of the LOs to search for semantic relations amongst them. Semantic relations are extracted because the educational content in our database is created by different experts over time. A customized text mining pipeline is utilized to extract the main topics covered each LO and compare them to topics covered by other LOs in the graph \cite{abu-rasheed_building_2023}. A relation is created between the two LOs if the semantic similarity is above a predefined threshold. Relations extracted between LOs are used to enhance the coverage and composition of context, which is provided to the LLM for generating an explanation of the connected materials in the learning recommendation.

The recommendation algorithm we use is a graph exploration and path weighing algorithm \cite{abu-rasheed_pedagogically-informed_2023}, based on Markov decision process (MDP) to identify an optimal learning path for each learner. Recommended learning path is explained in terms of its content selection (why each node is recommended), as well as its relation to the learning goal (how a node supports achieving the desired learning goal).
To explain the recommended path, we extract four main types of information from the KG, see Fig.1: 

\begin{figure}
  \centering
  \includegraphics[width=\linewidth]{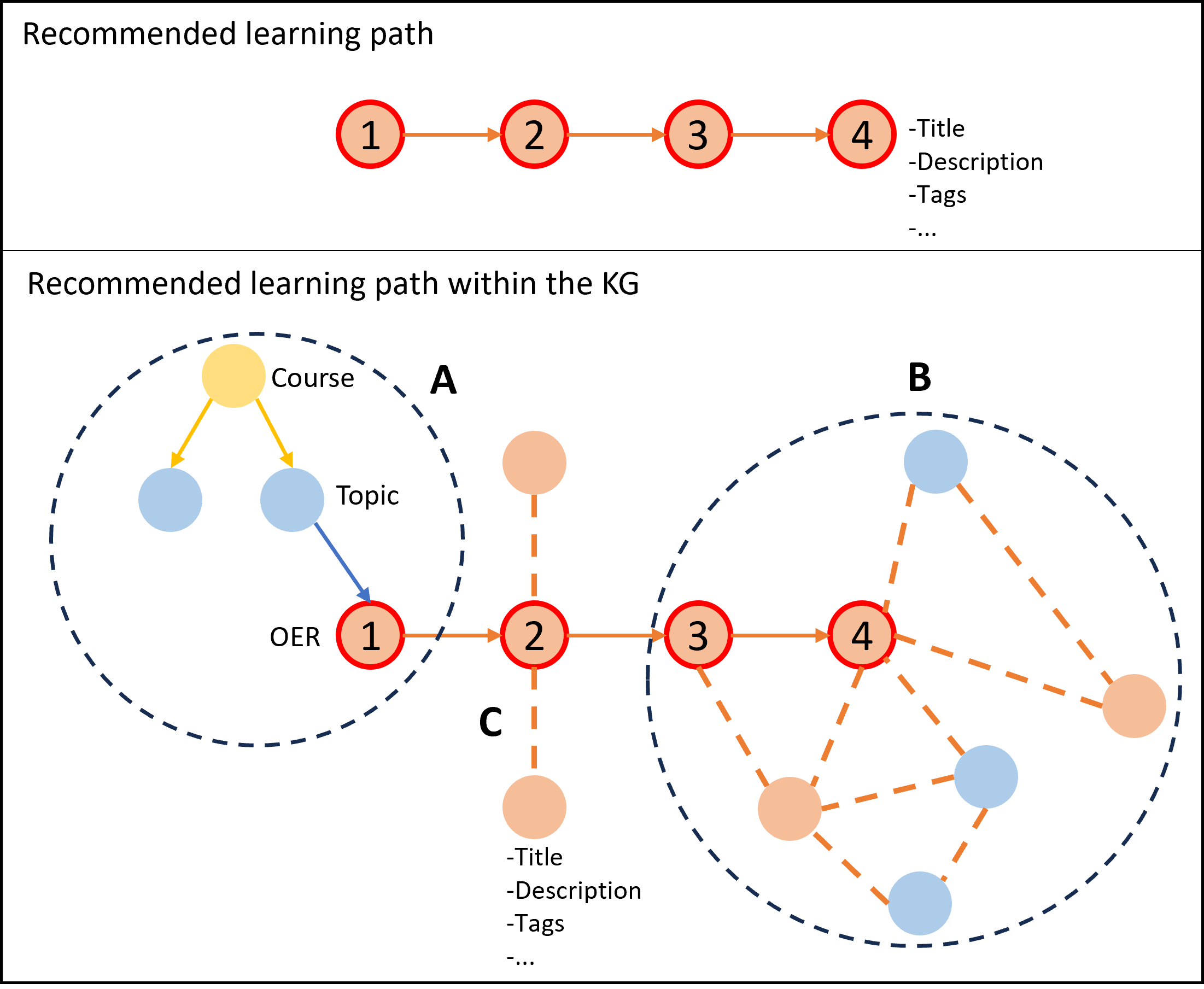}
  \caption{Structural information added to the LLM context from the KG. Top: learning path as an output of the recommendation system. Bottom: recommended path as it appears in the KG. Area (A): hierarchical structure of the learning goal. Area (B) KG community around LO3 and LO4. Connection (C): semantic relation extracted by the relation extraction algorithm.}
\end{figure}

\begin{enumerate}
    \item \textit{the hierarchical structure of the LO based on the pre-defined taxonomy}: this structure informs the LLM about the placement of a recommended LO within its original curriculum, which is defined by the human content-curator of the corresponding learning goal, see Fig. 1 area (A).
    \item \textit{Semantic relations to similar LOs in the KG}: which is discovered by the relation extraction algorithm based on the textual content and the semantic similarity between the LOs, see Fig. 1 connection (C).
    \item \textit{KG communities around LOs}: : which are areas on densely connected LOs in the KG. These communities reflect in our KG fields of application, in which several LOs are usually required together as a group, see Fig. 1 area (B).
    \item \textit{Supporting metadata from connected LOs in the KG}: where not only the relations to semantically similar LOs are utilized, but also the textual content of the related LOs’ metadata, which sheds an additional light on the context of the recommended LO.
\end{enumerate}

Using this additional information from the KG, we construct the GPT-4 prompt on the task level, as well as on the context one.

\subsection{GPT-4 Prompt and Explanation Construction}
The extended explanations for recommended learning paths are generated using the “GPT4-1106-preview” model, which we will refer to here as GPT-4 for simplicity. This model is the most capable model offered by OpenAI at the time of conducting this research, with training data that is updated till April 2023. To use the model, we utilize OpenAI’s API, which allows parsing the prompt to the GPT-4 model with additional data, information, and instructions, in order to guide the model output generation. We design our pomp to include a main body and a contextual part, see Fig. 2. Prompt’s body represents the direct query from the user. It is constructed as a set of tasks that the model ha to perform. This is to enhance the focus of the query and ensure less deviation from the user intent. 

The contextual part of the prompt is extracted directly from the KG, based on a search strategy that integrates the four types of information discussed in 3.1, into the prompt’s context. The context is designed to include: 1) The role that the model assumes when answering a user’s query, e.g., answering as a teacher. 2) Required definitions that are needed to understand the terminology, especially in the cases where the domain-specific meaning of a word differs from its general meaning. 3) The information from semantically connected LOs in the KG, which is provided to the model in the form of “supporting content” in the query’s context. The model’s role and terminology definitions are also influenced by a direct input from the domain experts, who can set the pedagogical and domain related guidelines for the model.

The response of the GPT-4 model is used to fill the information gaps in an explanation template. The design of explanations for the learning recommendations is tightly connected to the pedagogical value of the explanation, to ensure an improved learning outcome. For that reason, each template includes specific information spaces or slots, which are filled by the LLM, guided by the data of the KG. It is important to mention here that parts of the explanation template can also be filled with a direct input from the experts. In this paper, however, we limit the scope of information in the explanation to that generated automatically by the GPT-4 model, in order to evaluate its performance against a golden standard, which is extracted from the expert’s input. 

\begin{figure}
  \centering
  \includegraphics[width=\linewidth]{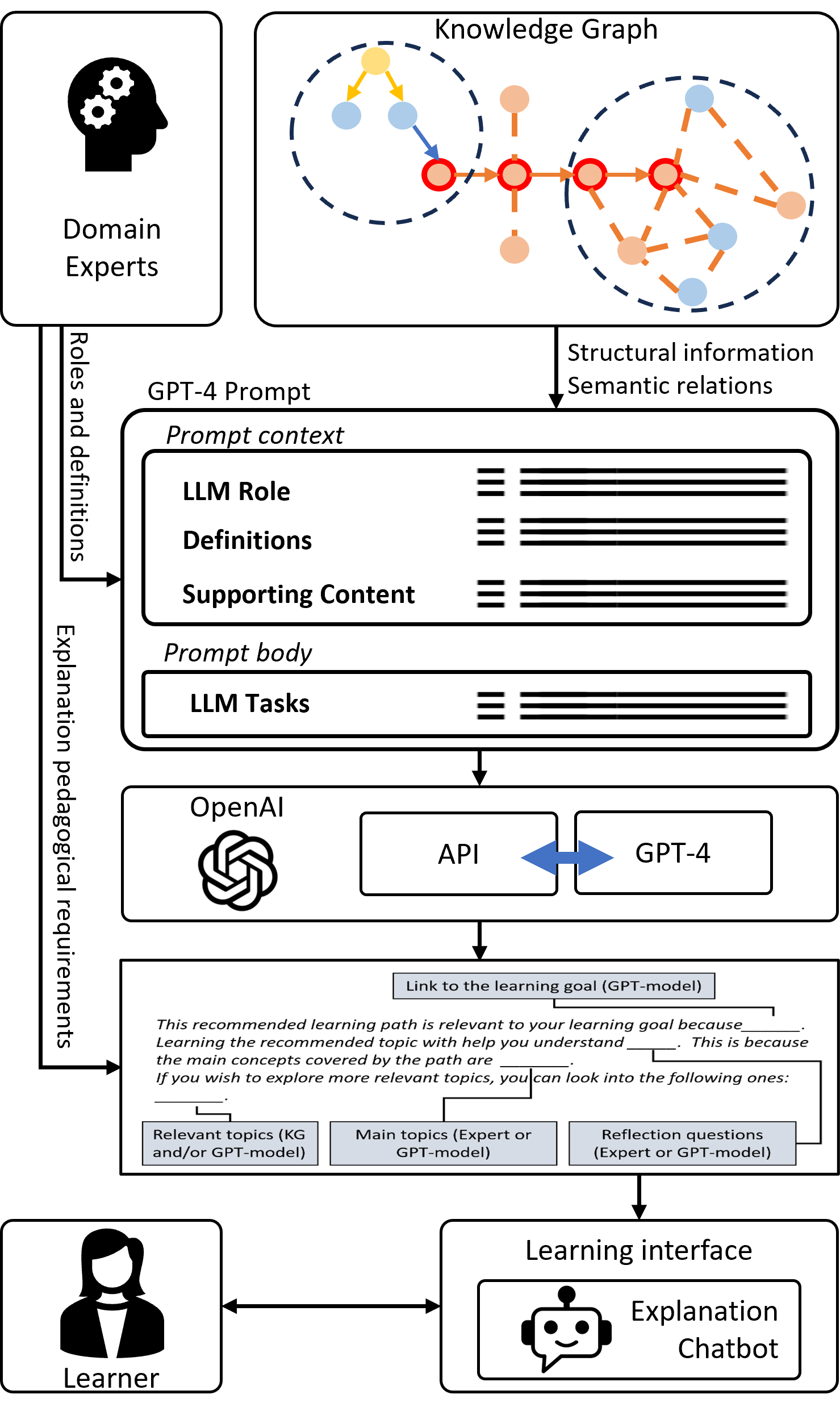}
  \caption{Proposed approach for constructing the GPT-4 prompt, with KG-based contextualization, as well as the Chatbot-based user interaction, and the expert roles in the design for context and explanation-templates.}
\end{figure}

\subsection{Chatbot-based interaction with the explanations}
Once the explanation is complete, it is provided to the user through an interactive interface element, which represents the chat features that LLMs offer. We refer to this approach as “conversational explainability”, which differs from regular textual and visual explanations by its ability to provide the explanation through a multi-step interaction with the user, taking into account previous chat messages to influence the following model output. The ability of GPT models to respond to individual questions, as well as to engage in a multi-step chat with the user, means that explanations can be extended with additional information based on user inputs and requests. This, however, would induce the risk of deviating from the main point of the explanations. Therefore, we limit the chatting feature of GPT-4 in this paper to a first response with confirmation. This approach stands for a one-shot query that is supported by a confirmation step from the model, to ensure that the chatbot understands the user request correctly. With this approach, a learner has the option to ask about the learning material itself, i.e., its content or why it is relevant for their learning goal, as well as the relationships between the learning materials are in the recommended path.

\section{Evaluation and Results}
To evaluate our proposed approach, we devise a hybrid quantitative/qualitative evaluation strategy. The quantitative evaluation utilizes Rouge-based measures, namely Rouge-N, Rouge-L and Rouge-Lsum. Qualitative evaluation is conducted through a questionnaire-based user feedback on the explanation approach and is outcomes. It involved domain experts and learners, who were asked to evaluate the generated explanations and their relevance to the learning materials and learning goal.

Through Rouge metric, we aim to measure the amount of text in the generated explanation, which offers precise information, against the amount of irrelevant text, which is either wrong, or simply a filler text that is task-related but has low relevance to the goal of the explanation. 

\begin{figure*}
  \centering
  \includegraphics[width=\textwidth]{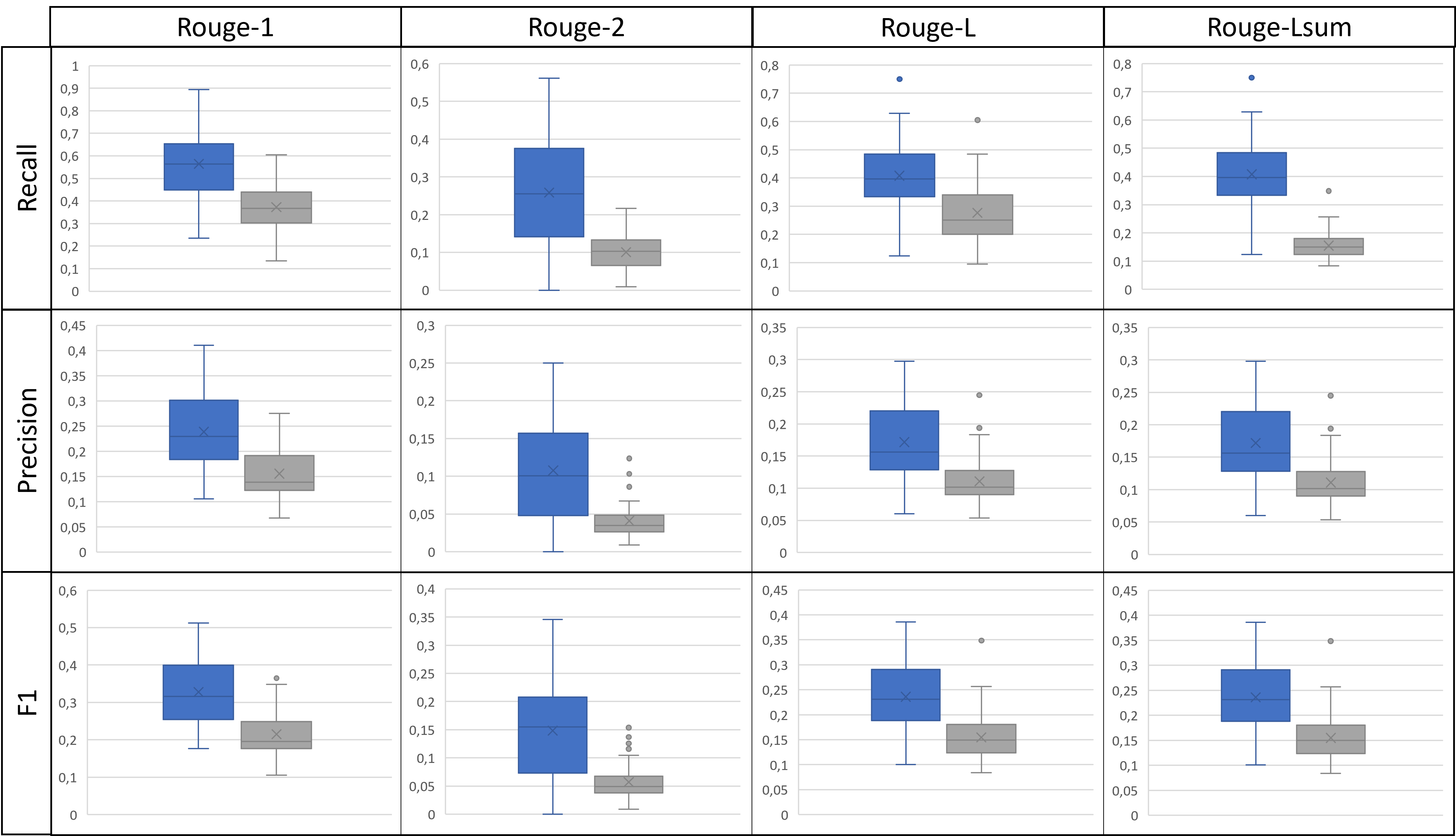}
  \caption{Recall, precision, and f1-measure values of the Rouge metric, for both explanation types: 1) with KG-based contextualization (blue), and 2) without contextualization (gray).}
\end{figure*}

\subsection{Experiment Set-up}
Rouge (Recall-Oriented Understudy for Gisting Evaluation) \cite{lin_rouge_2004} is an evaluation metric commonly used in natural language processing (NLP) to assess the quality of automated text summarization and machine translation systems. It employs both recall and precision to compare the overlap of n-grams (contiguous sequences of words) between model-generated summaries and human-generated ones. We utilize this metric to measure the overlap between reference explanations from a human text and each of the explanations generated automatically by the GPT-4 model, i.e., with a KG-based contextualization, and without it. To ensure a fair comparison between the two explanation approaches, we fix the length of the text generated by the GPT-4 model. This is because Rouge measures take into consideration the total amount of n-grams in the model generated text. 

The reference text is composed from human defined descriptions and reflection information in the LO’s metadata. In this paper, we do not account for the phrasing of the explanation, i.e., sentence structures and writing styles, since there is no certain phrasing or writing style that can be considered as “correct” or “referential”. Instead, we focus on the word choice and word patterns that appear in the generated text and match the patterns in the reference text.

\subsection{Dataset}
To calculate Rouge values, we construct a reference data set of reference explanations. 52 samples of human-generated texts were collected from clarifications of the recommended content of 10 different learning-path recommendations. For each reference explanation, we generate two candidate explanations, one through the proposed KG-based contextualization approach, and one without the context.

\subsection{Results and Discussion}
We calculate the recall, precision, and f1-measure from four Rouge measures: Rouge-1, which is the measure used to for the overlap between single words, i.e., unigrams, Rouge-2 that accounts for two-word patterns, i.e., bigrams, Rouge-L that takes into account the longest matching patterns between the reference text and the candidate one, and Rouge-Lsum which is a variant of Rouge-L that calculates the overlap on the sentence level, not the complete text. Calculated values of the different measures, see Fig. 3, show a clear enhancement of the precision, recall, and thus f1-measure of the explanations generated with KG-based context. The results also show that recall scores are best with Rouge-Lsum measure, while the precision and f1-measure values are better with Rouge-1. With a domain expert support, we evaluate that none of the automatic explanation samples was wrong or misleading, which leads to the conclusion that the amount of less irrelevant text in the non-contextualized explanations was considerably higher than the one in the contextualized explanations, since both have comparable lengths.

Qualitatively, the explanations were provided in a user study to a group of eight learners and five domain experts. Participants included two PhD holders, four PhD candidates and seven graduate students. We surveyed the participants for their perception on the explanation quality, and the overall impressions, remarks, and limitations of the two explanation approaches. The survey evaluation revealed an enhanced acceptance of the explanations when it is generated with KG-based context. On a Likert scale, the quality of the contextualized explanation reached 4.7/5. The amount of irrelevant text, which was correct in general but not related to the user intent from the explanation, was also reported to be less in the contextualized explanations, which aligns with quantitative results. Participants also highlighted important limitations that the LLM revealed when generation the explanation. Three out of five domain experts pointed out that not only the word selection, but also the phrasing of the explanation plays an important role in determining its meaning. Four experts also emphasized that the explanation is not only a task of answering a user question, but also includes high-level reflection information, which enables the learner to understand how the learning content can affect their own context, e.g., their daily work. This level of reflection exceeds the ability of the LLM and requires a human mentor to provide it based on their understanding of the learner’s context.

This feedback shows a limitation in our valuation approach, which should also include the effect of the explanation phrasing to the comparison with reference explanations. The nature of LLMs and the complexity of evaluating what a correct phrasing is, challenges this evaluation, and defines at the same time several prospects for continuing this on-going research. Other limitations that we address in our study include: 1) The limited sample sizes in the experiment, which is used to provide a proof-of-concept and set the starting point for a larger scale test. 2) The comparison between GPT-4 and other LLMs is required to measure LLM-based performance deviations. 3) We do not include user-specific date in our contextualization approach, to comply with the GDPR, since GPT-4 is provided by a third party. A local LLM is intended to be used for further evaluating the role of user data in enhancing the relevance of the explanation to user contexts, as well as materials contexts.

\section{Conclusion}
In this paper, we proposed an evaluated an approach for utilizing KGs as a source of contextual information that supports LLMs in generating more relevant explanations of learning recommendations. Our approach extracts different types of information from the KG and constructs a contextual part of a GPT-4 model’s prompt accordingly. The design of the prompt’s context is conducted with experts-in-the-loop to ensure the fulfillment of the pedagogical requirements from the explanation. Experts also participated in the design of the final explanation shape and content, which was offered to the learners as a textual template filled by the GPT-4 model. A chatbot-based interaction with the user is used to provide the explanations as answers to user questions. We evaluate the proposed approach qualitatively and quantitatively using Rouge measures. Evaluation results provided a proof of the proposed concept, showing a reduced amount of less-relevant text in the generated explanation. The results of the evaluation also pointed out limitations of the LLM’s performance which defines a starting point for future work to further evaluate the role of the explanation phrasing and the use of user-data for further personalization of the explanation text.

\bibliography{sample-2col}

\end{document}